\documentclass[letterpaper, 10 pt, conference]{ieeeconf}  

\IEEEoverridecommandlockouts                              

\usepackage{graphicx} 
\usepackage{subfigure}
\usepackage{amsmath} 
\usepackage{amssymb}  
\usepackage{multirow}

\title{\LARGE \bf
Robust Human Identity Anonymization using Pose Estimation
}

\author{Hengyuan Zhang$^{*}$, Jing-Yan Liao$^{*}$, David Paz, Henrik I. Christensen
\thanks{$^{*}$These members contributed equally to this publication.}
\thanks{Affiliation: Contextual Robotics Institute,
        University of California, San Diego, 9500 Gilman Dr, La Jolla, CA 92093}
\thanks{\copyright 2022 IEEE.  Personal use of this material is permitted.  Permission from IEEE must be obtained for all other uses, in any current or future media, including reprinting/republishing this material for advertising or promotional purposes, creating new collective works, for resale or redistribution to servers or lists, or reuse of any copyrighted component of this work in other works.}
}

\begin{document}

\maketitle
\thispagestyle{empty}
\pagestyle{empty}

\begin{abstract}
Many outdoor autonomous mobile platforms require more human identity anonymized data to power their data-driven algorithms. The human identity anonymization should be robust so that less manual intervention is needed, which remains a challenge for current face detection and anonymization systems. In this paper, we propose to use the skeleton generated from the state-of-the-art human pose estimation model to help localize human heads. We develop criteria to evaluate the performance and compare it with the face detection approach. We demonstrate that the proposed algorithm can reduce missed faces and thus better protect the identity information for the pedestrians. We also develop a confidence-based fusion method to further improve the performance.

\end{abstract}

\section{INTRODUCTION}

Outdoor mobile robots have the huge potential to benefit human life. Applications such as autonomous driving or delivery are coming to our life.
A critical step before they can be deployed at scale is the pedestrian motion prediction problem, given that human motion is highly uncertain and multi-modal. To address this problem, recent algorithms rely on data-driven approaches. However, collecting these data poses risks of leaking identity information. Areas having identity information, such as human faces and license plates are required to be anonymized by recent regulations. Recent released datasets follow this approach to blur faces and license plates.

Face identity anonymization consists of two parts, face detection and applying an anonymization algorithm to the detected region. Face detection can be achieved by applying a standard object detector trained on a face specific dataset. Face anonymization methods can then be applied on facial regions; examples include Gaussian blur and more recent methods such as DeepPrivacy \cite{application_learning_anonymization_deep_privacy}. In this work, we are majorly concerned about the face detection phase, where we localize the identity sensitive region. 

Over these years of research and development, face detection has evolved from using hand-crafted features, such as Haar, to deep-learning ones. Although there are various techniques to tackle this problem, the recent trend in face detection algorithms is treating it as a sub-problem in the object detection field \cite{application_learning_face_detection_Zhu2020TinaFaceSB}. At the time of writing this paper, the state-of-the-art face detection method is YOLO5Face \cite{application_learning_face_detection_YOLO5Face_qi_2021}, which is based on the YOLOv5 \cite{yolov5_repository} object detector.

Nevertheless, the robustness associated with these methods is subject to their ability to detect facial features across various orientations and small regions; this can drastically vary across various image resolutions. Instead, we propose to infer the human head from the skeleton generated by pose estimation algorithm. Given that human keypoints across the entire body will cover a much larger region than the face alone, this allows us to anonymize identity information more robustly.

This paper is organized as follows. We review the prior work in face detection, face anonymization, and keypoint detection in section \ref{related_work}. Our proposed method is introduced in detail in section \ref{method}. We evaluate the proposed method and discuss the metrics in section \ref{experiments}. Finally, we conclude with a number of key takeaways in section \ref{conclusion}.

Our major contributions are:

1. We propose to leverage keypoint detectors to infer head location, which increases the identity sensitive region detection range and robustness, thus better protecting the identity of pedestrians.

2. We propose a metric to evaluate the proposed method and compare it fairly with the face detectors in the anonymization context.

3. Furthermore, we show that a confidence-based fusion method can further improve the performance.

\begin{figure*}
    \centering
    \includegraphics[scale=1.1]{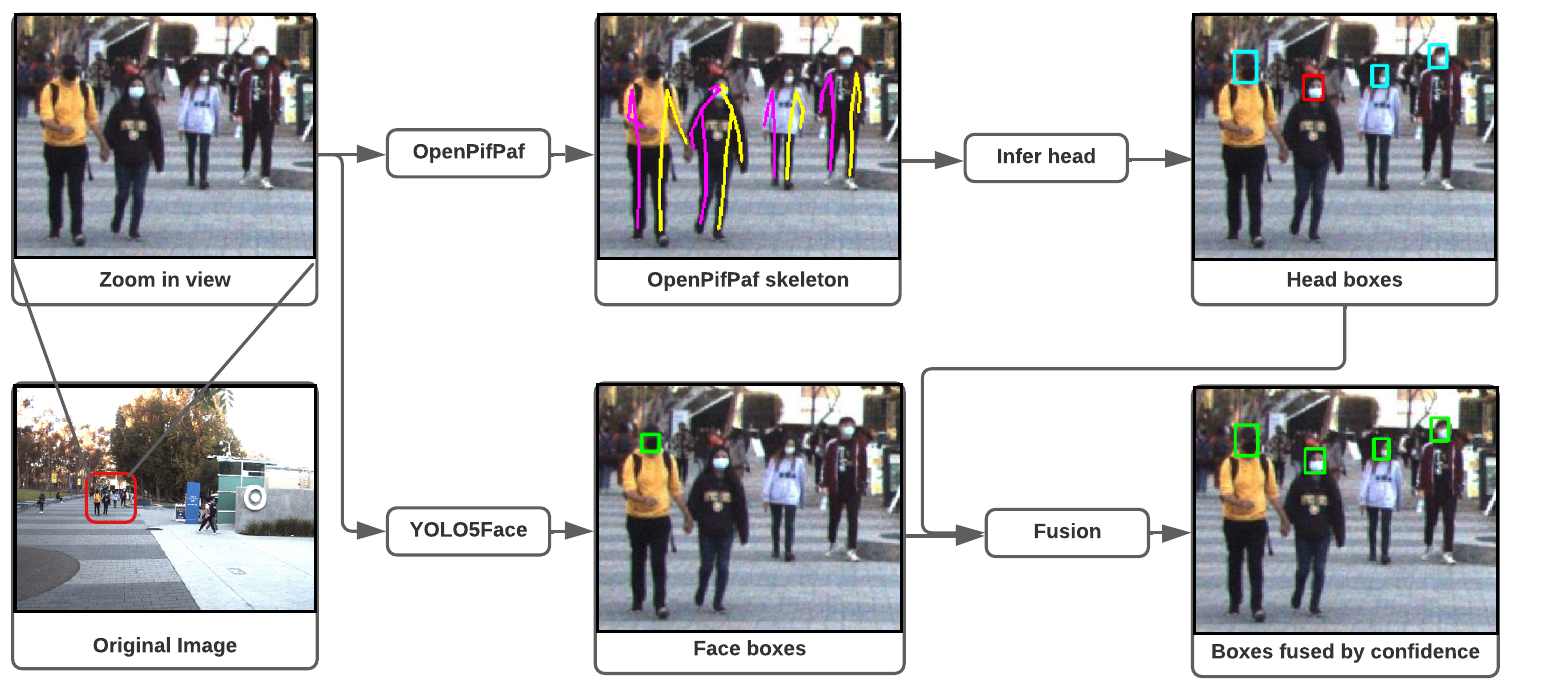}
    \caption{We infer head bounding box from the skeleton predicted by a pose estimation algorithm (OpenPifPaf), and detect face bounding boxes use a face detector (YOLO5Face). These head and face boxes are fused to produce the final output. (Results are presented in a zoom in view of the original image)}
    \label{fig:pipeline}
\end{figure*}

\section{RELATED WORK}\label{related_work}

\textbf{Face Detection:} As the first step of the anonymization system, face detection can directly impact the performance and robustness. There are several approaches for face detection. Before deep learning based methods were utilized, handcrafted features such as Haar \cite{Haar} were used to detect faces. Recently, after the benchmark dataset WiderFace \cite{dataset_widerface_yang_CVPR_2016} was released, face detection grew rapidly and focused on challenges such as multi-scale. To tackle these challenges, many works leverage the knowledge from general object detectors in the face detection task. FDNet \cite{fdnet_Zhang2018FaceDU}, as an example of two-stage detectors, employed multi-scale training, multi-scale testing with light-head Faster RCNN \cite{FasterRCNN} and redesigned anchors to help multi-scale performance. For single-stage methods, TinaFace \cite{application_learning_face_detection_Zhu2020TinaFaceSB}, which is based on RetinaNet \cite{retinanet_focal_loss_Lin_2017_ICCV}, introduced the Inception Module to enhance capability of multi-scale detection, and used DIoU \cite{DIoU} as regression loss for small face detection. The state of the art, YOLO5Face \cite{application_learning_face_detection_YOLO5Face_qi_2021}, modified the YOLOv5 \cite{yolov5_repository} bottleneck part in the architecture to increase robustness for large and small faces. Based on SSD \cite{SSD}, PyramidBox \cite{application_learning_face_detection_Tang2018PyramidBoxAC} utilized contextual information with the proposed Pyramid Anchors (PA), the Low-level Feature Pyramid Network (LFPN), and the Context-sensitive Prediction Module (CPM) to predict multi-scale faces. 

\textbf{Face Anonymization:} In identity anonymization systems, once faces are detected, an anonymization algorithm can be applied to the detected region. Even though this is not the focus of our paper, we briefly discuss the prior work to provide the necessary context. This can be done easily by blocking the region with random pixels, blurring, or pixelation. However, recent datasets targeting outdoor tasks not only need to anonymize the identity, but also need to reduce its impact on pedestrian detection. Recent approaches use learning to address this problem. Instead of trying to hide the identity by destroying the details, the methods focus on changing the appearance, either adding distortion or generating a completely new face. For example, DeepPrivacy \cite{application_learning_anonymization_deep_privacy} uses a generative adversarial network to produce faces that matches the background context but also looks completely different from the original face. CIAGAN \cite{application_learning_anonymization_CIAGAN_Maximov_2020_CVPR} further brings in control of features of anonymization as well as temporal consistency in a video sequence.

\textbf{Pose Estimation:} There are two approaches in pose estimation: top-down and bottom-up methods. The former methods could be perceived as a person detector with a single-person pose estimator, while the latter predicts body joints individually and associates them afterwards. By leveraging a large amount of data and methods from human detection, top-down methods can achieve competitive results in human pose estimation. For example, Mask R-CNN \cite{application_learning_mask_rcnn_he2017ICCV} extends instance segmentation to pose estimation by predicting a one-hot map for each joint. However, top-down methods are not suitable for our case because partial occlusions can lead to missed detections in the human detection stage. This is the result from its two-stage detecting nature. For instance, the performance of human detection could be unstable if only partial observations are presented in the image frame, making it difficult to detect skeletons properly, thus posing challenges to localize head poses. In contrast, this can be addressed by bottom-up methods in most cases. Moreover, bottom-up methods are also more suitable for real-time applications as they are faster \cite{BENGAMRA2021104282}. The pioneer work, DeepCut \cite{application_learning_pose_estimation_DeepCut_Pishchulin_CVPR_2016}, detects keypoints individually and associates detected joints with an integer linear program, which is computationally expensive. Later works accelerate prediction time with greedy decoders, such as OpenPose \cite{application_learning_pose_estimation_OpenPose_CoRR_2018}, PersonLab \cite{application_learning_pose_estimation_PersonLab_George_CoRR_2018}, and our method of choice, OpenPifPaf \cite{application_learning_pose_estimation_OpenPifPaf_Kreiss_2021}. OpenPifPaf handles pose in various scale by predicting the joint location and size. Although top-down methods generally handle scale variance better than bottom-up methods, we find that OpenPifPaf satisfies our requirements as the human faces that are too small don't need to be detected for anonymization.

\section{METHOD}\label{method}

To overcome the limitations of the face-detection based strategies, we propose to use human body pose estimation methods to infer head position. The pipeline is shown in Fig. \ref{fig:pipeline}. In this section, we introduce the keypoint detection and association method we use, OpenPifPaf. We then outline the process of estimating head position from the keypoints generated by OpenPifPaf. Finally, we discuss the tracking and fusion method that jointly leverages face and head detections.

\subsection{OpenPifPaf}

Our method is based on the human body pose estimation model, OpenPifPaf. OpenPifPaf consists of three parts, backbone feature extraction, Composite Intensity Fields (CIF) and Composite Association Fields (CAF). 

The backbone stage is extract feature from image with ResNet \cite{application_deep_learning_resnet_HeCVPR2016} or ShuffleNetV2 \cite{Ma2018ShuffleNetVP}, and the output feature would be shared by CIF and CAF. The CIF is a 1x1 convolution that predicts semantic keypoints, which is identical to the Part Intensity Field (PIF) introduced in \cite{pifpaf_cvpr_2019}. For human subjects, the keypoints correspond to body parts such as joints. For a specific part, the network outputs the confidence score, its two dimensional vector from each pixel and a scale. Given that the feature maps estimated by the CIF prediction head are coarse, a convolution is applied over the coarse targets using a bivariate Gaussian kernel, which yields a high resolution confidence map. 

The CAF part shares the same backbone; however, this head predicts a confidence, two vectors that point towards the two parts this association connects and two spreads for spatial precision of the regression and two joint sizes.

The output of OpenPifPaf is a set of keypoint locations and the confidence. Such keypoints are meaningful parts of a human, such as left shoulder, right eye and nose. Given the predefined connectivity, we can link the key points and produce the skeleton of a human. 

We note that OpenPifPaf generalizes well on our data without fine tuning. The detection range of it is much better than the YOLO5Face detector. It is also more robust on the same range, which raises our interests to examine its use for face detection and identity anonymization.

\subsection{Head Prediction}

Given the human body pose predictions generated by OpenPifPaf, which consist of skeletons for the COCO \cite{dataset_Lin2014MicrosoftCOCO} dataset, we focus on inferring head positions for anonymization purposes. The actual procedure differs depending on whether facial keypoints are predicted or not. 

When there are facial keypoints available from OpenPifPaf, we can readily estimate the head center from these facial keypoints. Then the bounding box dimensions (height and width) of the head are inferred from torso length, which is given by the distance from the shoulder keypoint to the hip keypoint. We assume a fixed ratio between the head dimension to the torso length in terms of pixels. This actually varies from person to person but serves as a good approximation for anonymization purpose. 

When facial keypoints are not available but shoulder and hip keypoints are available, the head position needs to be inferred. This is done simply by centering a bounding box horizontally between the shoulder keypoints, and vertically on top of the shoulder keypoints plus a neck length. Again, the bounding box dimensions are inferred from the torso length by assuming a fix ratio between them. We also assume the neck length has a fixed ratio with respect to the dimension of the head and experimentally find it using our data.

\subsection{Head and Face Fusion}

Given the head boxes predicted by OpenPifPaf, it is intuitive to consider fusion of these head boxes with the face boxes generated by the YOLO5Face.

Since our goal is to reduce missed faces as much as possible to better protect identity, we design our approach to be more tolerant to false positives. Thus the simplest fusion strategy involves keeping both results. However, we can also remove face boxes that are covered by the head boxes since the region will be anonymized.

A more risky strategy is to remove a head bounding box if there exists a face prediction within. This can potentially increase the probability of exposing identity information; however, this strategy is justified by the fact that usually when a face box is associated with a head box, the subject in question is closer to us and we are confident about the face location.

Finally, we can filter the bounding box by its confidence. We keep the bounding box with a higher confidence score if for a head bounding box there exists a face prediction within. The confidence score for face bounding box comes directly from the output of YOLO5Face while the confidence score for a head bounding box is the average of the keypoint confidence from OpenPifPaf.

\subsection{Head and Face Tracking}

Given the predicted head bounding boxes from OpenPifPaf, we leverage a tracking pipeline to reduce false positives but also to incorporate temporal consistency. A Kalman Filter \cite{kalman_filter_1960} was applied with association based on center-distance. We modify the SORT \cite{sort_tracking_bewley} tracker for our application. Initially we associated detections with existing tracks based on IOU and GIoU \cite{GIoU_Rezatofighi_2018_CVPR}; however, this often generated poor associations as faces are often too small and may not overlap between adjacent image frames. Thus, we choose an $l^2$ center distance based association which gives better performance.

\section{EXPERIMENTS}\label{experiments}

We evaluate the proposed algorithm and present the results in this section. To this end, we label two video sequences, one for parameter tuning, which consist of 257 frames, and another 544 frames in different scene for testing. Then, we define a metric that allows fair comparison. Finally, we present the results of our algorithm and its comparison with the face detection algorithms. It is noteworthy that the data was recorded during COVID and masks can be observed but the neural networks were not fine-tuned for this specific case.

\subsection{Metric and Labeling}

\begin{figure}
    \centering
    \includegraphics[scale=0.6]{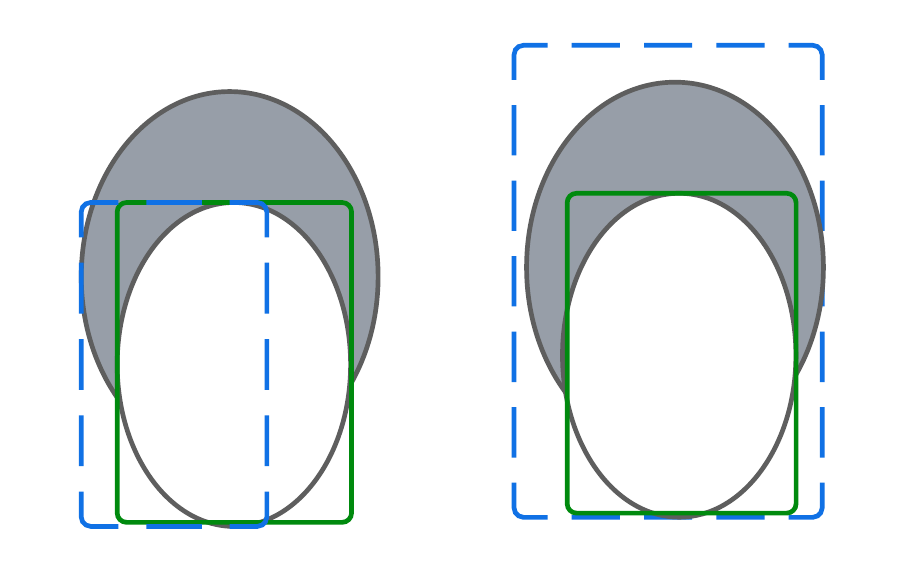}
    \caption{Green solid line boxes are face labels and blue dashed line boxes are detections. Left and right detection both have 0.5 IoU with the face label.}
    \label{fig:metric_examples}
\end{figure}

A fair comparison of our approach against the face detectors is challenging. Standard evaluations use intersection over union (IOU). But considering that we focus on identity anonymization, this can be achieved by blurring most of the face and there will be no consequences of including other parts of the head. For example, two examples with an IoU score of 0.5 are shown in Fig. \ref{fig:metric_examples}. While both achieve the same score, the example on the right evidently better protects the sensitive region compared to the example on the left. Even though additional regions are also included, applying anonymization to the region has lower risk. 

Another challenging point is the nature of the algorithms. Face detectors capture the faces and not the head as depicted by the green bounding boxes, The bounding boxes may change largely when the head facing angle changes. Since our approach mostly captures the bounding boxes of head, which includes hair and is more invariant to the facing direction, evaluating them using a standard IoU metric would lead to inconsistent comparisons.

\begin{figure}
    \centering
    \includegraphics[scale=0.6]{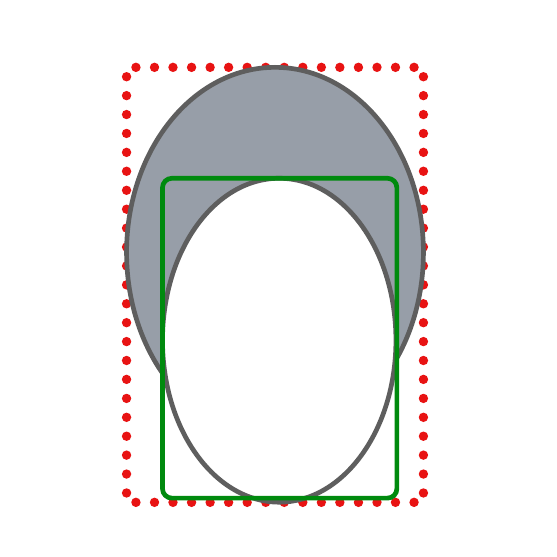}
    \caption{The green solid line box is a face label and the red dotted line box is a head label.}
    \label{fig:two_labels}
\end{figure}

Therefore we use a two label strategy, as shown in Fig. \ref{fig:two_labels}. We label both faces (green solid line) and heads (red dotted line). We then denote the bounding box for a detection by $D$, a face label by $F$, and a head label as $H$.
Moreover, two criteria for evaluation are introduced. First, the face criterion is defined as $$\frac{D \cap F}{F} > \alpha,$$
which implies that at least a portion $\alpha$ of the face label should be included in $D$; thus most of the identity information can be anonymized.

Second, the head criterion is defined by $$\frac{D \cap H}{D} > \beta,$$
which implies that at least a portion $\beta$ of the detection should be included in the head label; thus the background portion within predicted bounding box is bounded.

We collected two video sequences in dense pedestrian walkways in a university area. For our labeled video sequences, both heads and faces of the pedestrians are labeled. Face labels are considered mostly the front of the head with skin, not including the ears and hair. Head labels include the hair. We provide labels up to the distance where human annotator believes that the face is potentially recognizable.
The video sequence for parameter tuning consists of 259 frames, with 1018 face labels and 1726 head labels. The video sequence for testing consists of 543 frames, with 1949 face labels and 3819 head labels.

\subsection{Comparison with Face Detection} \label{method_comparison}


\begin{table*}[]
    \vspace{2mm}
    \centering
    \begin{tabular}{|c|c|cccc|cc|c|c|}
    \hline
& Label & \multicolumn{4}{c|}{Face and Head} & \multicolumn{2}{c|}{Head} & FP count & FPS \\
    \hline 
Task & Match & Both & Face & Head &
None & Head & None & & \\
\hline
\multirow{2}{*}{Detection} &
YOLO5Face & 1637 & 2 & 26 & 247 & 252 & 1455 & 551 & 18.1\\
& OpenPifPaf Head & 1772 & 5 & 28 & 107 & 1417 & 290 & 1540 & 5.4 \\
\hline
\multirow{4}{*}{Detection Fusion}
& keep head & 1838 & 5 & 18 & 51 & 1463 & 244 & 2545 & \multirow{4}{*}{4.04}\\
& keep face & 1829 & 4 & 28 & 51 & 1465 & 242 & 2543 &\\
& keep both & 1840 & 3 & 18 & 51 & 1465 & 242 & 3784 &\\
& by confidence & 1839 & 4 & 18 & 51 & 1463 & 244 & 2545 &\\

\hline
\multirow{2}{*}{Tracking} & YOLO5Face & 1468 & 5 & 58 & 381 & 197 & 1510 & 473 & 18.1 \\
& OpenPifPaf Head & 1654 & 9 & 38 & 211 & 1300 & 407 & 1288 & 5.4\\
\hline
    \end{tabular}
    \caption{The result of matching predicted OpenPifPaf head boxes and YOLO5Face face boxes with the groundtruth head labels and face labels (details in section \ref{method_comparison}).}
    \label{table:detection_two_label_result}
\end{table*}

\begin{table*}[]
    \centering
    \begin{tabular}{|c|c|c|cccc|cc|c|}
    \hline
& & Label & \multicolumn{4}{c|}{Face and Head} & \multicolumn{2}{c|}{Head} & FP count\\
    \hline 
& Task & Match & Both & Face & Head &
None & Head & None & \\
\hline
\multirow{3}{*}{Unfiltered} &
\multirow{2}{*}{Detection} &
YOLO5Face & 1637 & 2 & 26 & 247 & 252 & 1455 & 551 \\
& & OpenPifPaf Head & 1772 & 5 & 28 & 107 & 1417 & 290 & 1540 \\
\cline{2-10}
&
{Detection Fusion}
& by confidence & 1839 & 4 & 18 & 51 & 1463 & 244 & 2545 \\
\hline
\multirow{3}{*}{Filtered} &
\multirow{2}{*}{Detection} &
YOLO5Face & 1634 & 2 & 26 & 234 & 249 & 1445 & 557 \\
& & OpenPifPaf & 1728 & 5 & 26 & 137 & 1368 & 326 & 644 \\
\cline{2-10}
& 
{Detection Fusion}
& by confidence & 1818 & 4 & 18 & 56 & 1443 & 251 & 1634 \\
\hline
    \end{tabular}
    \caption{Comparison of results with boxes filtered by size for both head labels and head prediction.}
    \label{table:filtered_results}
\end{table*}

Based on the criteria proposed in the previous section, we evaluate the proposed OpenPifPaf based algorithm, the face detector and the fusion method on our testing video sequence. 

The evaluation process are as follows. First, we associate each face label with a head label, which is based on the assumption that each face label should be within a head label. Then for the pedestrians, we have a group of front facing individuals that have both face labels and head labels, and another group only have head labels (back of the pedestrians). Second, we associate the detections with the head labels using the Hungarian algorithm \cite{theory_hungarian_Kuhn1955TheHM}. Finally, we evaluate the predicted boxes with the labels using the proposed criteria. Unless otherwise specified, we set $\alpha$ and $\beta$ to 0.5.

For the targets that have a face label and a head label, we check if the associated detection box satisfies both of the criteria, face only, head only or none. We denote the result as match both, face, head or none. For the targets that has only a head label, we only evaluate the head criterion for the associated detection. We denote the result as match head or none. If a detected box is not associated with any of the labels or doesn't satisfy any of the criteria with the associated labels, it is counted as a false positive case.

As shown in the Table \ref{table:detection_two_label_result}, OpenPifPaf predicted detections produce significantly more face matches compared to YOLO5Face; however, they also generate considerably more false positives. The results suggest that OpenPifPaf predicted boxes are doing better in terms of finding face regions. Furthermore, given that we set a higher cost to missing a face than generating false positives, we consider the performance of the OpenPifPaf based method acceptable.  

The fusion method further improves the performance. If we keep both the face and the head boxes, the approach produces the lowest number of missed faces while also generating the highest number of false positives. Applying other fusion methods when a predicted face from YOLO5Face is in the predicted head from OpenPifPaf allows us to greatly reduce false positives while the number of missed faces only drops slightly. Among these fusion methods, fusion by confidence and the keep head method produce similar results, both are slightly better than the keep face method. Post processing shows that the similarity of the two fusion methods is because OpenPifPaf in general gives higher confidence scores compared to the scores from YOLO5Face, even when the predicted boxes are less accurate. To make the confidence fusion more effective, an extension to our work is to tune the confidence scores from the two neural networks using the same dataset.

\begin{figure}
    \centering
    \includegraphics[scale=0.40]{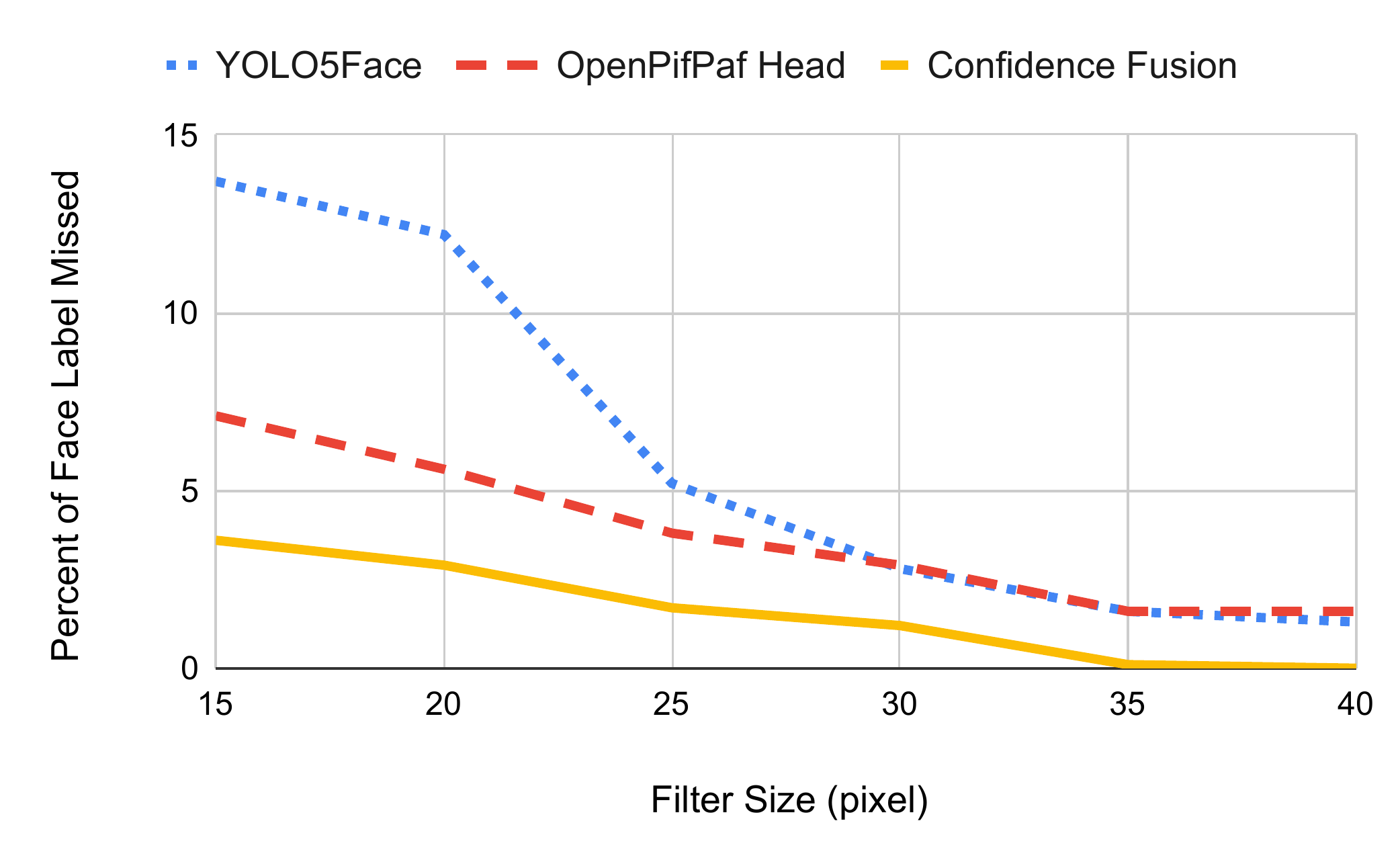}
    \caption{Percent of face missed for different methods when filter size changes.}
    \label{fig:face_missing_rate}
\end{figure}

Another advantage is that the OpenPifPaf's predicted boxes miss fewer facial regions when the faces are large. We verify this by removing head labels whose maximum dimension is smaller than a certain threshold. Then we pick those face labels that has a head label associated with it and check how much are these labels not detected by the predicted boxes. We consider it missed when the face criterion is not satisfied. As shown in Fig. \ref{fig:face_missing_rate}, The fusion results significantly reduce the rate of missed faces for various bounding box size thresholds and OpenPifPaf predicted boxes does better for smaller boxes. The fusion method achieves zero missing rate for heads whose maximum dimension are larger than 40 pixels.

\begin{figure}
    \centering
    \includegraphics[scale=1.02]{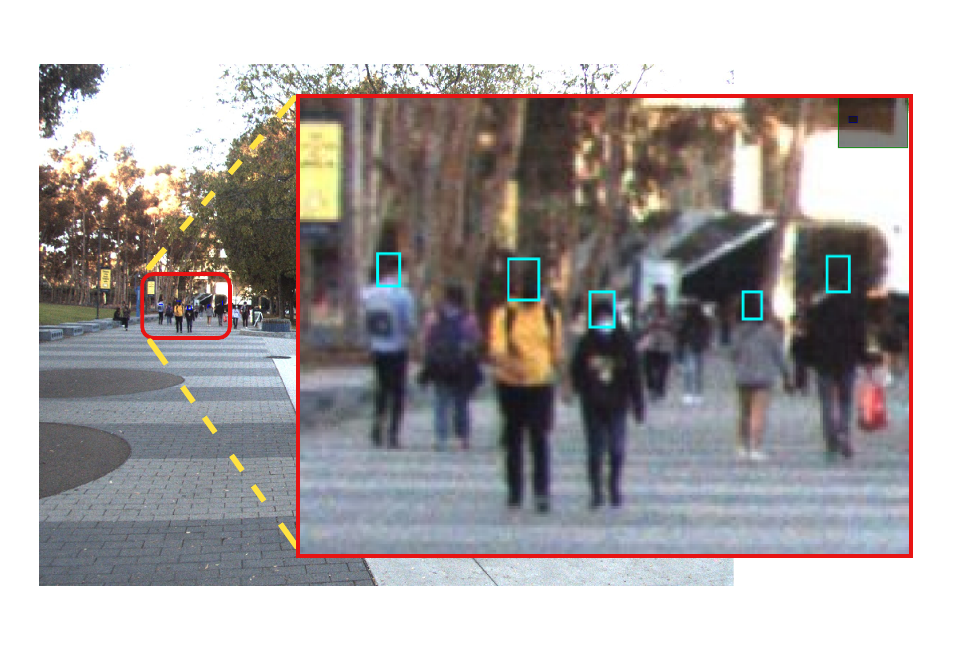}
    \caption{OpenPifPaf based method can begin to find head boxes earlier than it is recognizable.}
    \label{fig:detect_early}
\end{figure}

In terms of the high false positive count, we would like to argue that this does not reflect the true performance of the algorithm. 

This is supported by our observation in the data. For example, as shown in Fig. \ref{fig:detect_early}. OpenPifPaf based method was able to infer head boxes (cyan bounding boxes) even when the pedestrians are very far away. These faces are so blurry that we did not find it necessary to provide labels for them, since we only provide labels up to the distance where human annotator believes that the face is potentially recognizable. These predictions are counted as false positive because we don't provide labels for them and it does no harm to anonymize them any way. This is not the general case where we consider a false positive in a random place, which might cause loss of information in other region if we anonymize it. 

We also show this by the following experiment. As shown in Table \ref{table:filtered_results}, we compare the result with or without filtering the label and predicted head boxes by its maximum dimension. We set the threshold to 15, which is considered a very small head size in an image of dimension 1920x1440. But with the threshold, we see that the correct cases dropped slightly while the false positive cases dropped significantly. It verifies our observation that OpenPifPaf was able to detect heads beyond the recognition range, where there are no labels. For this reason, the resulting high false positive count of OpenPifPaf based results does not imply that it is not a good algorithm for anonymization. 

Another interesting findings from the Table \ref{table:detection_two_label_result} is that the tracking pipeline decreases performance. And the performance decreases more for the YOLO5Face. From our experience working with the labels, this is mostly caused by the head movement when human are walking. Our head movement is not linear, but coupled with ups and downs for each step. This does not align with the linear velocity assumption for the Kalman Filter. Given that the YOLO5Face predicts smaller boxes, it is more vulnerable to the error cause by this. Another reason is the nonlinear motion from the data collection platform. We collected the data using two cameras mounted on top of a vehicle. Thus the target motion is also coupled with the motion of ego vehicle.

\begin{figure}
    \centering
    \includegraphics[scale=1.2]{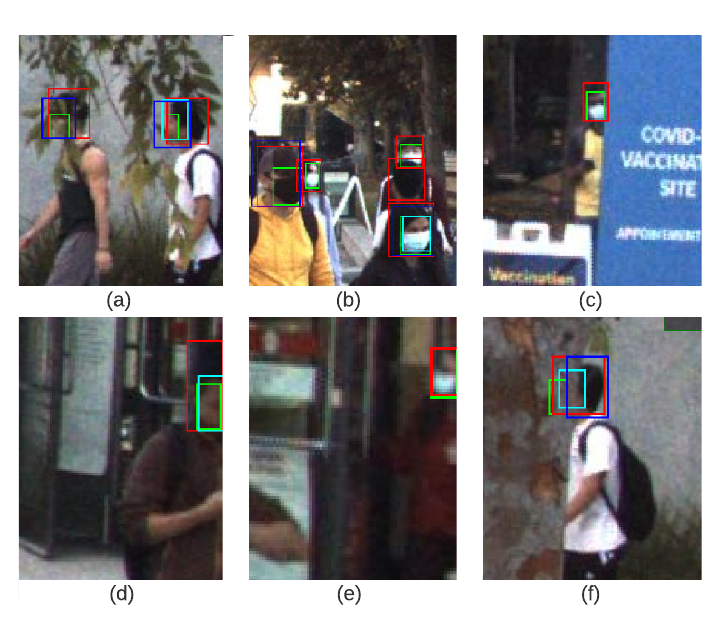}
    \caption{Example results of YOLO5Face and the OpenPifPaf based method in challenging scenarios such as occluded (a,b,c), truncated (d,e) and hidden (f) faces. Bounding boxes represent face labels (green), head labels (red), YOLO5Face detections (cyan) and the OpenPifPaf based method detections (blue).}
    \label{fig:detect_examples}
\end{figure}

We compare the OpenPifPaf based method with YOLO5Face in challenging cases where the faces are occluded, truncated or hidden. We present some results in Fig. \ref{fig:detect_examples}. In Fig. \ref{fig:detect_examples}(a), the person on the right was detected by both methods while the person of the left was only detected by the proposed method. This suggests that the OpenPifPaf based method can handle face occlusion better than YOLO5Face. But it also requires a significant part of the body to present. This is verified by Fig. \ref{fig:detect_examples}(b)(c) where multiple faces are missed by both methods. Similarly, the OpenPifPaf based method is not good at handling truncated faces, as shown in Fig. \ref{fig:detect_examples}(d)(e), since part of the body is also outside of the image. YOLO5Face fails at Fig. \ref{fig:detect_examples}(d) but succeeds in Fig. \ref{fig:detect_examples}(e). Fig. \ref{fig:detect_examples}(f) presents an example where the face was considered hidden but both methods were able to make the detection. It is also worth noting that in Fig. \ref{fig:detect_examples}(b), there is a face with mask only detected by OpenPifPaf based method. This is one of the benefits of using the proposed method, being more robust to facial coverings and decorations.

\subsection{Result Change for Different Threshold}

In the following experiments, we would like to show how different face criterion threshold $\alpha$ and head criterion threshold $\beta$ would influence the result. 

During these experiments, we set one threshold to 0.5 and increase the other threshold from 0.1 to 0.9 by a step size 0.1. For all pedestrians that have both face label and label, we classify them into match both, face, head or none and count the total number for each category. The results are presented in figures.

\begin{figure}
    \centering
    \includegraphics[scale=0.40]{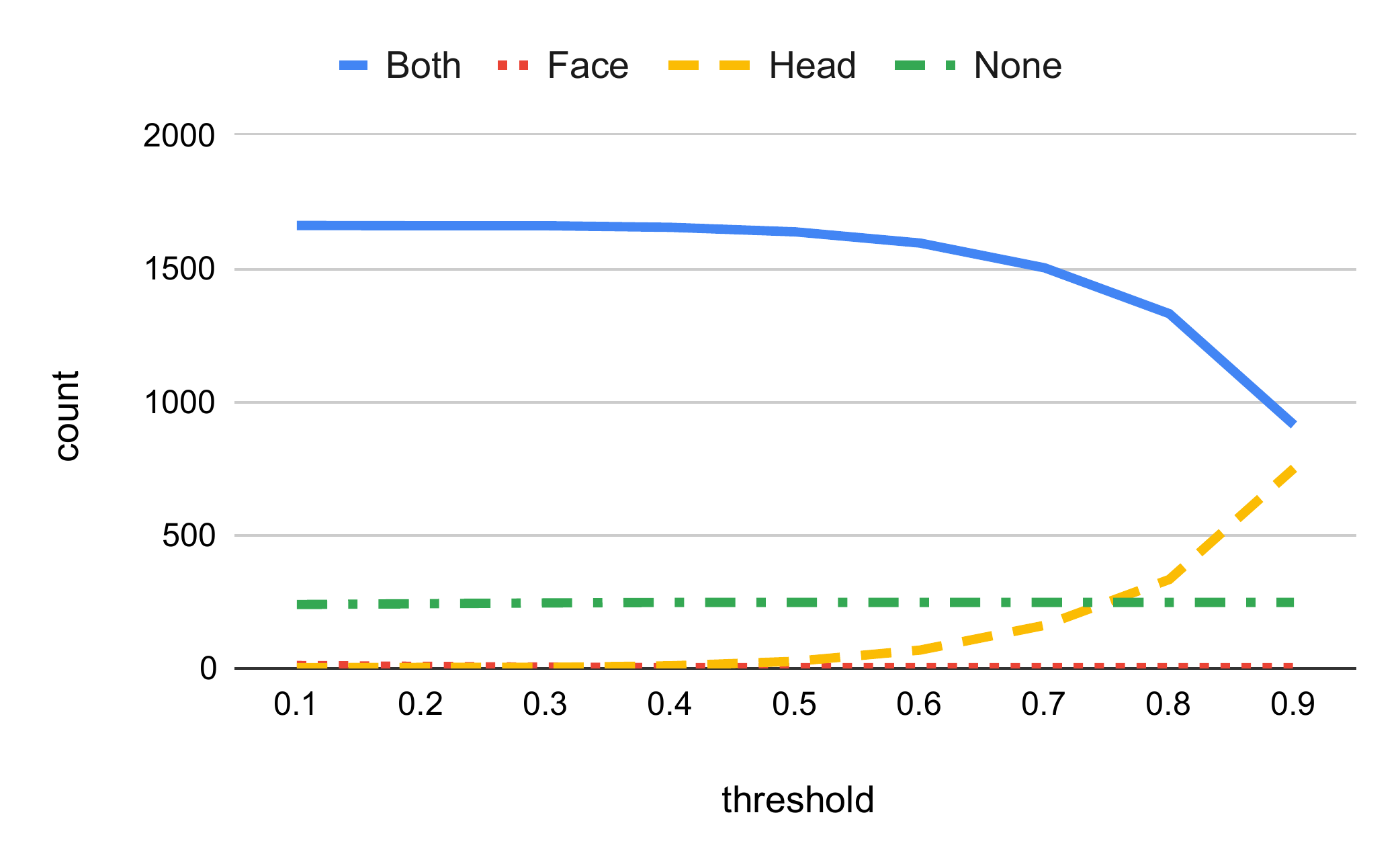}
    \caption{YOLO5Face result on different face label  threshold.}
    \label{fig:face_threshold_yolov5face}
\end{figure}

\begin{figure}
    \centering
    \includegraphics[scale=0.40]{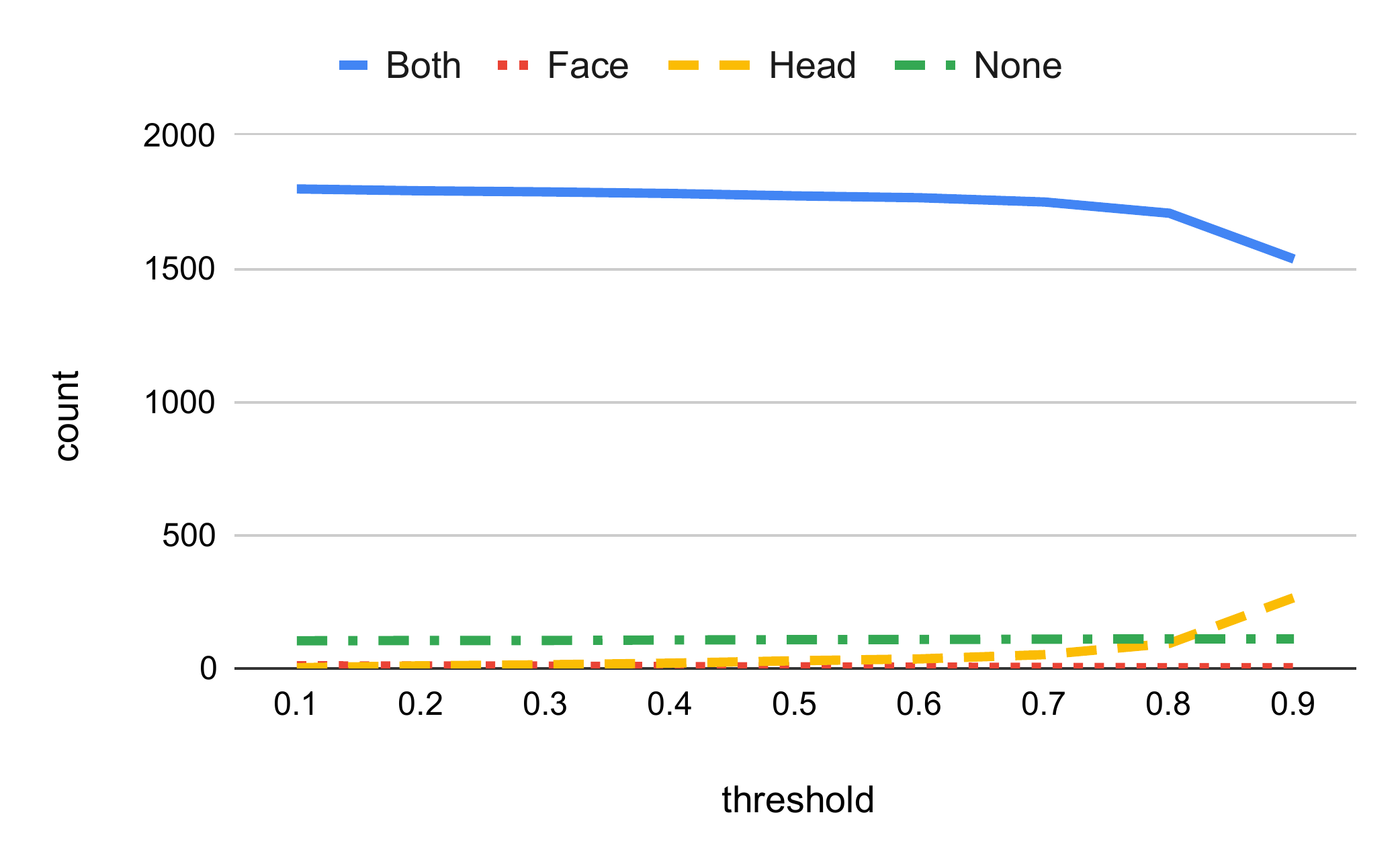}
    \caption{OpenPifPaf predicted head result on different face label threshold.}
    \label{fig:face_threshold_openpifpaf}
\end{figure}

For face criterion threshold $\alpha$, the higher it is, the more facial region we require the predicted bounding box to cover. It can be seen from Fig.  \ref{fig:face_threshold_yolov5face} that when $\alpha$ is higher than 0.5, the performance of YOLO5Face drops quickly. This is not the case for OpenPifPaf predicted boxes as shown in Fig. \ref{fig:face_threshold_openpifpaf}. The result is very natural as OpenPifPaf predicted larger boxes, which is likely to cover the face box even when the matching criterion becomes more strict. This implies that using OpenPifPaf predicted box would cover most of the face information. Thus better protect the identity.

\begin{figure}
    \centering
    \includegraphics[scale=0.40]{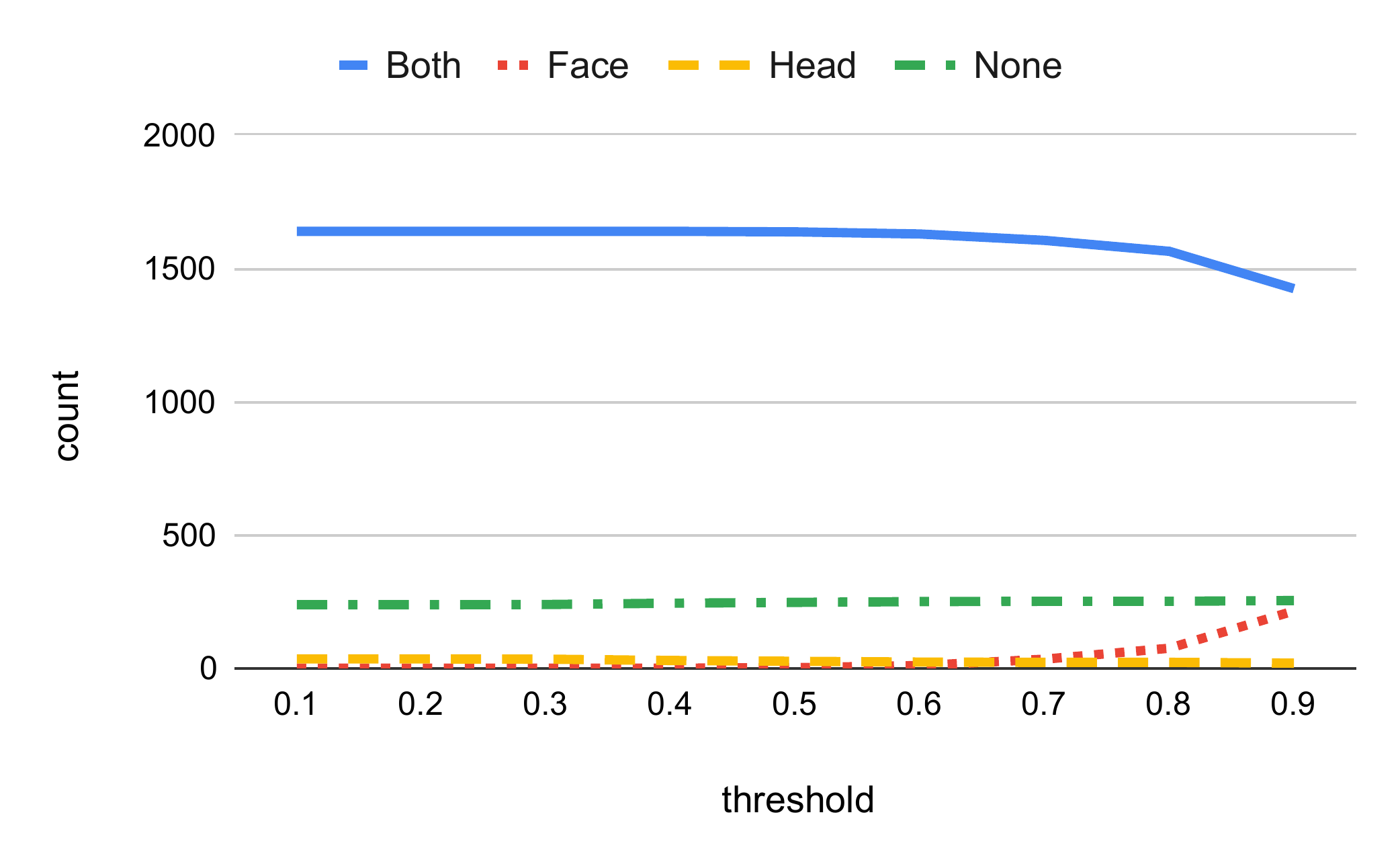}
    \caption{YOLO5Face result on different head label threshold.}
    \label{fig:head_threshold_yolov5face}
\end{figure}

For head criterion, YOLO5Face is less influenced by the change of its threshold $\beta$ as shown in Fig. \ref{fig:head_threshold_yolov5face}. Comparatively, OpenPifPaf predicted boxes drops performance significantly when the criterion becomes stricter. But the overall accuracy is still high when the threshold is set to 0.6.

\begin{figure}
    \centering
    \includegraphics[scale=0.40]{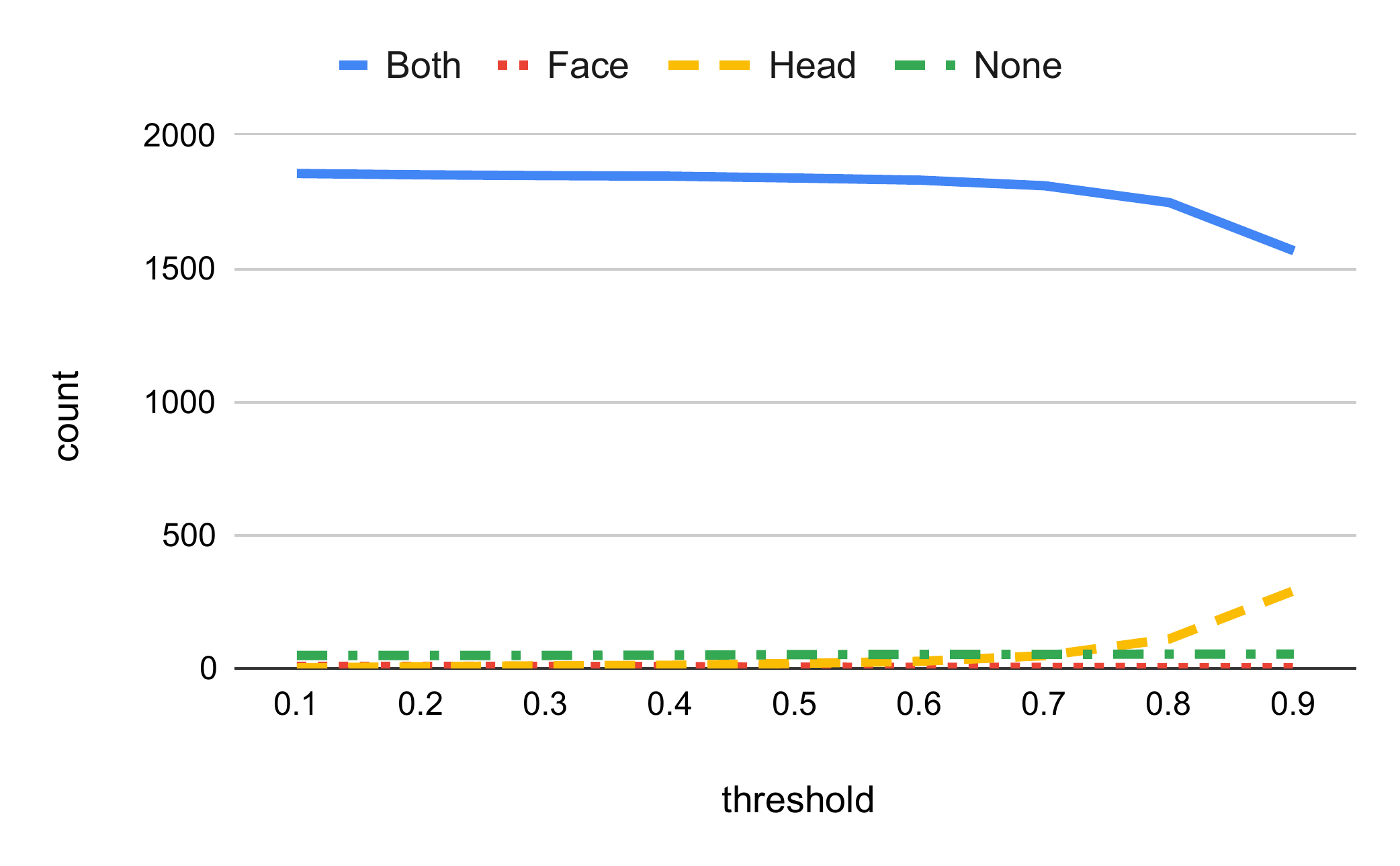}
    \caption{Fusion by confidence result on different face label threshold.}
    \label{fig:face_threshold_fusion}
\end{figure}

\begin{figure}
    \centering
    \includegraphics[scale=0.40]{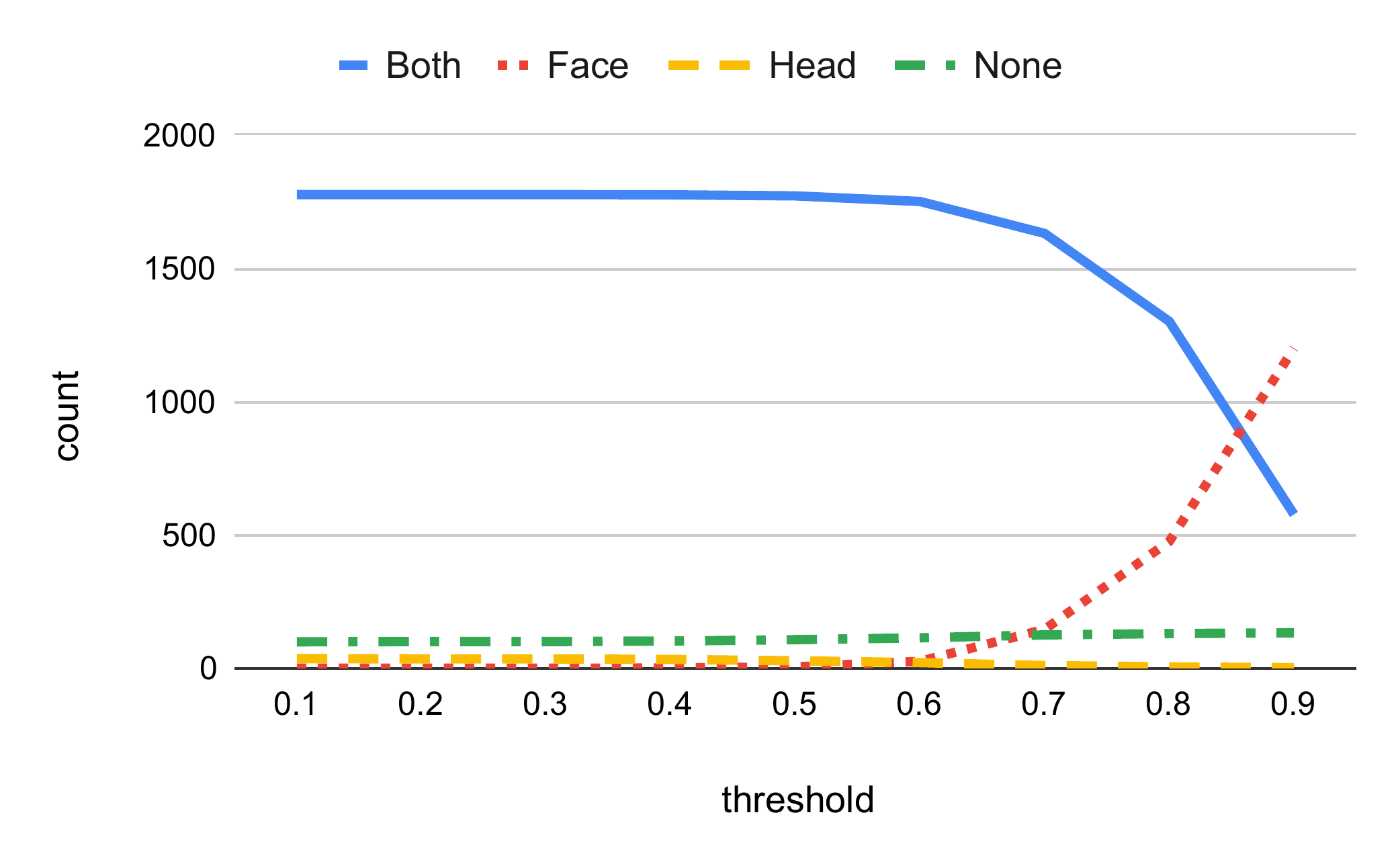}
    \caption{OpenPifPaf result on different head label threshold.}
    \label{fig:head_threshold_openpifpaf}
\end{figure}

\begin{figure}
    \centering
    \includegraphics[scale=0.40]{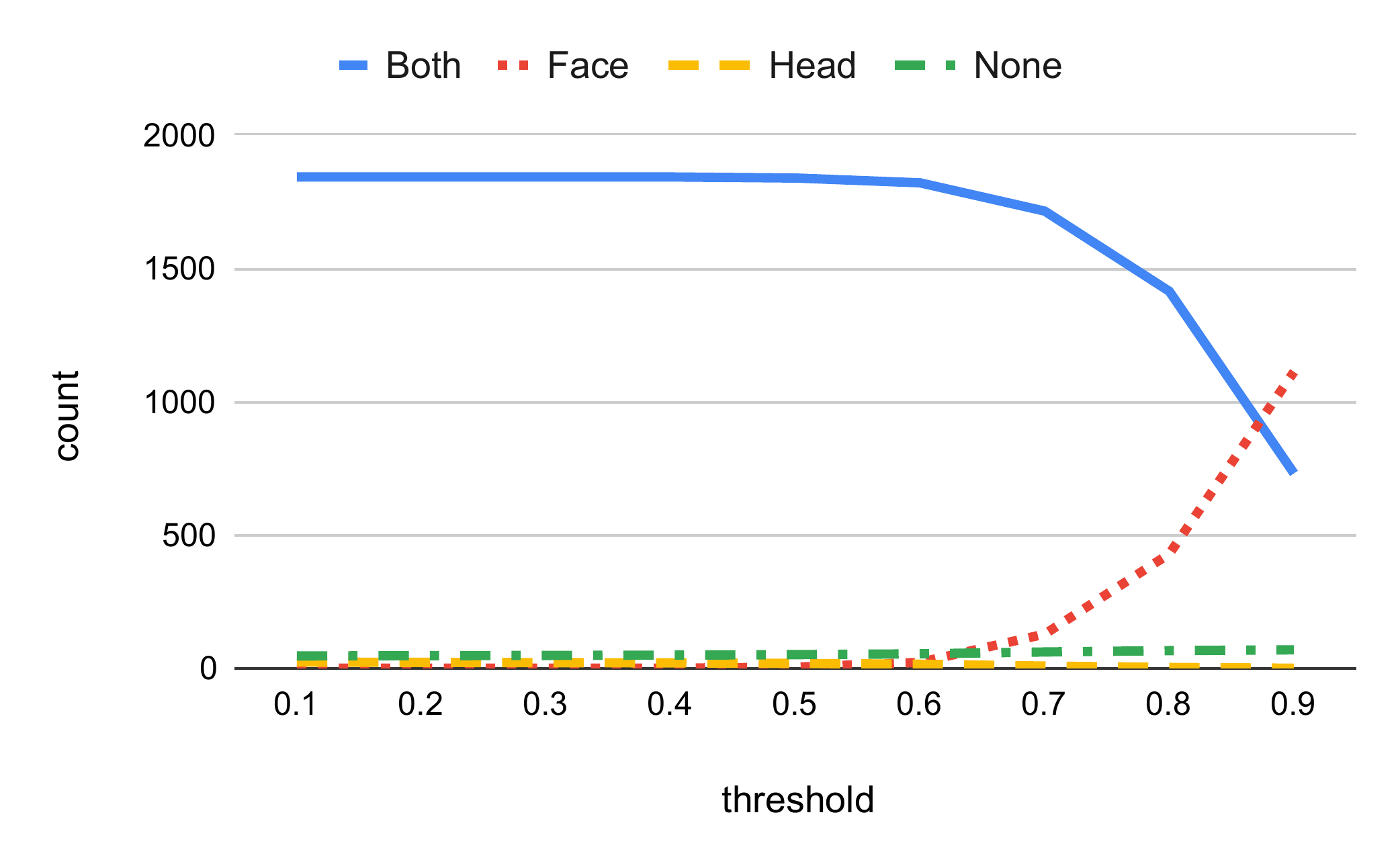}
    \caption{Fusion by confidence result on different head label threshold.}
    \label{fig:head_threshold_fusion}
\end{figure}

\subsection{Inference Speed}
Lastly, we would like to discuss the performance trade-off for our anonymization pipeline. We report frame rate for various methods testing on single 2080ti. As shown in the Table \ref{table:detection_two_label_result}, the YOLO5Face model runs at 18.1 frames per second (FPS), which is considered real-time in autonomous driving settings where many datasets are recorded at 10Hz. When combined with Openpifpaf, the proposed pipeline runs at 4 FPS, which no longer satisfies the real-time requirement. 

While the proposed method performs better at the expense of computational costs, we see its value in offline use cases and potential to reduce its run time. As an offline method, the proposed algorithm with higher accuracy can reduce the risk of identity leak in large-scale datasets. It will also require less post-examination that can save human efforts. Moreover, our proposed algorithm is a plug-and-play pipeline, which means as the state-of-the-art model in pose estimation evolves, the system could be faster and more accurate.

\section{CONCLUSION}\label{conclusion}

We proposed to use the skeleton from pose estimation algorithm to infer head bounding boxes for identity anonymization. Based on this, we further fuse the output with face detector output to get better results. We show that the proposed methods can significantly reduce the failure cases compare to the face detectors alone. Thus we believe this can be apply to a face anonymization system to better protect the identity. For future work, we will further improve the head prediction localization accuracy and develop more sophisticated fusion method.

\bibliographystyle{unsrt}
\bibliography{citation}

\end{document}